\documentclass{svproc}
\usepackage{graphicx}
\begin{document}
\mainmatter              % start of a contribution
\title{Comparative study of Transformer and LSTM Network with attention mechanism on Image Captioning}
%
% \titlerunning{Hamiltonian Mechanics}  % abbreviated title (for running head)
%                                     also used for the TOC unless
%                                     \toctitle is used
%
\author{Pranav Dandwate\inst{1} \ Chaitanya Shahane\inst{2} \ Vandana Jagtap\inst{3} \ Shridevi C. Karande\inst{4}\
}
%
% \authorrunning{Ivar Ekeland et al.} 
% abbreviated author list (for running head)
%
%%%% list of authors for the TOC (use if author list has to be modified)
% \tocauthor{Ivar Ekeland, Roger Temam, Jeffrey Dean, David Grove,
% Craig Chambers, Kim B. Bruce, and Elisa Bertino}
% %
\institute{Dr.Vishwanath Karad MIT World Peace University,Pune,India\\
\email{https://mitwpu.edu.in/},\\ 
% WWW home page:
% \texttt{http://users/\homedir iekeland/web/welcome.html}
% \and
}

\maketitle              % typeset the title of the contribution

\begin{abstract}
In a globalized world at the present epoch of generative intelligence, most of the manual labour tasks are automated with increased efficiency. This can support businesses to save time and money. A crucial component of generative intelligence is the integration of vision and language. Consequently, image captioning become an intriguing area of research. There have been multiple attempts by the researchers to solve this problem with different deep learning 
architectures, although the accuracy has increased, but the results are still not up to standard.
\newline
This study buckles down to the comparison of Transformer and LSTM with attention block model on MS-COCO dataset, which is a standard dataset for image captioning. For both the models we have used pre-trained Inception-V3 CNN encoder for feature extraction of the images. 
\newline
The Bilingual Evaluation Understudy score (BLEU) is used to checked the accuracy of caption generated by both models. Along with the transformer and LSTM with attention block models,CLIP-diffusion model, M2-Transformer model and the X-Linear Attention model have been discussed with state of the art accuracy. 

% We would like to encourage you to list your keywords within
% the abstract section using the \keywords{...} command.
\keywords{Transformer, LSTM, Attention Mechanism, InceptionV3, MS-COCO }
\end{abstract}
\section{Introduction}
In this paper, two widely used architectures for image captioning are compared, drawing on recent key research in the disciplines of computer vision and natural language processing [9]. This task involves giving the description of the visual content of an image using natural language, so a system is needed that can both understand the image and generate grammatically correct statements. The objective of this project is to identify an effective pipeline for connecting the visual input and the output sentences in order to produce meaningful phrases from images. Image captioning is a complex task as the model must understand the image and prioritise certain aspects while generating a descriptive caption. Previous attempts with different architectures, such as RNN and LSTM, have faced the issue of vanishing gradients for longer sentences [1] [2]. In this research paper, we aim to evaluate the accuracy of image captioning using two well-known models, the Transformer [3] and the LSTM with attention mechanisms [10]. Furthermore, we will also delve into more advanced architectures for image captioning, including the Clip Diffusion model, as well as the Top-Down and Bottom-Up approaches to attention [5], which have demonstrated state-of-the-art accuracy in caption generation. By exploring these models, we aim to gain a deeper grasp of the best practices for image captioning and to contribute to the advancement of this field.
\newline
\newline
In the study, two machine learning models were evaluated under the same conditions using the MS-COCO dataset for training and validation. The MS-COCO dataset comprises everyday objects and people, with each image annotated with five captions to assist the model in identifying and describing the objects. Both models were trained on an AWS GPU[24] instance of type g4dn.xlarge.
\newline
\newline
The evaluation of generated captions employs various performance metrics, including BLEU, METEOR, ROUGE, CIDEr [8], and SPICE, as measurement indicators. BLEU [7]  and METEOR [6] originate from machine translation, while ROUGE is rooted in text abstraction. CIDEr and SPICE, on the other hand, are particular metrics tailored to image captioning.
% \begin{eqnarray*}
%   \dot{x}&=&JH' (t,x)\\
%   x(0) &=& x(T)
% \end{eqnarray*}
% with $H(t,\cdot)$ a convex function of $x$, going to $+\infty$ when
% $\left\|x\right\| \to \infty$.

%
\section{Related Work}
The task of Image Captioning requires the integration of both Computer Vision and Natural Language Processing techniques .In recent years, there has been a growing interest in this field, leading to the development of several methods for solving the problem
\newline
Before Deep Learning approaches for generating caption for images, there have been many probabilistic model including Multiple Instance Learning model, which detects the objects from the given image and generate the object words by applying CNN network. Image captions are used to train the generated words and through probability distribution most likely words are choose and caption is generated.
\newline
Graph based Image Captioning like G-Cap [26] showed 10\% higher accuracy than its previous modes. Idea behind this approach is to represent images, caption and regions as nodes and link them into a graph. This model did not focus on edge weight which could increase the captioning accuracy.
\newline
Markov-Logic Network(MLN) [25] uses probabilistic inference to estimate uncertainty in generated captions. Captions are passed,attention mechanism is used to add visual context and is passed to MLN. This approach generated captions close to human-generated captions. 
 \newline
 Initially, Image Captioning was tackled using Recurrent Neural Networks (RNNs) [2] and Long-Short Term Memory (LSTM) networks. Although these methods generated captions effectively, they faced the issue of vanishing gradients where the gradients decreased as the model processed longer sequences.  To address this challenge, researchers proposed the use of an Attention Mechanism in conjunction with RNNs or LSTMs, allowing the network to focus on the most important parts of the input and make predictions based on relevant information.
\newline
There have been multiple modifications on attention mechanisms including Soft attention[21] which focuses on specific parts of the input data when making the prediction. The word “soft” implies it uses continuous distribution to weigh the importance of the different parts of input data. On the contrary Hard attention selects one region at random and only focuses on that particular area.
Global attention [27] which considers the entire input data globally without giving preference to any specific part. It allows the model to learn from all parts of the input and make predictions based on a holistic understanding of the data. In Local Attention[27] prior to computing the attention weight in the windows to the left and right of the selected point for alignment, first determines a position for alignment. It then gives the context vector a weight.
\newline
Adaptive attention with visual sentinel[22] is a popular attention technique, in which the model chooses whether to concentrate on the picture or the sentinel. This method resolves the problem with earlier attention systems that had difficulty anticipating short words that the decoder could not see, such as "for" and "and."
More recently, the Transformer architecture [3]  has emerged as a leading approach for Image Captioning. This architecture, based on self-attention mechanisms, has proven successful in a variety of NLP tasks, including machine translation and text classification.

\begin{figure}
    \centering
        \includegraphics[width=6cm,height=7cm]{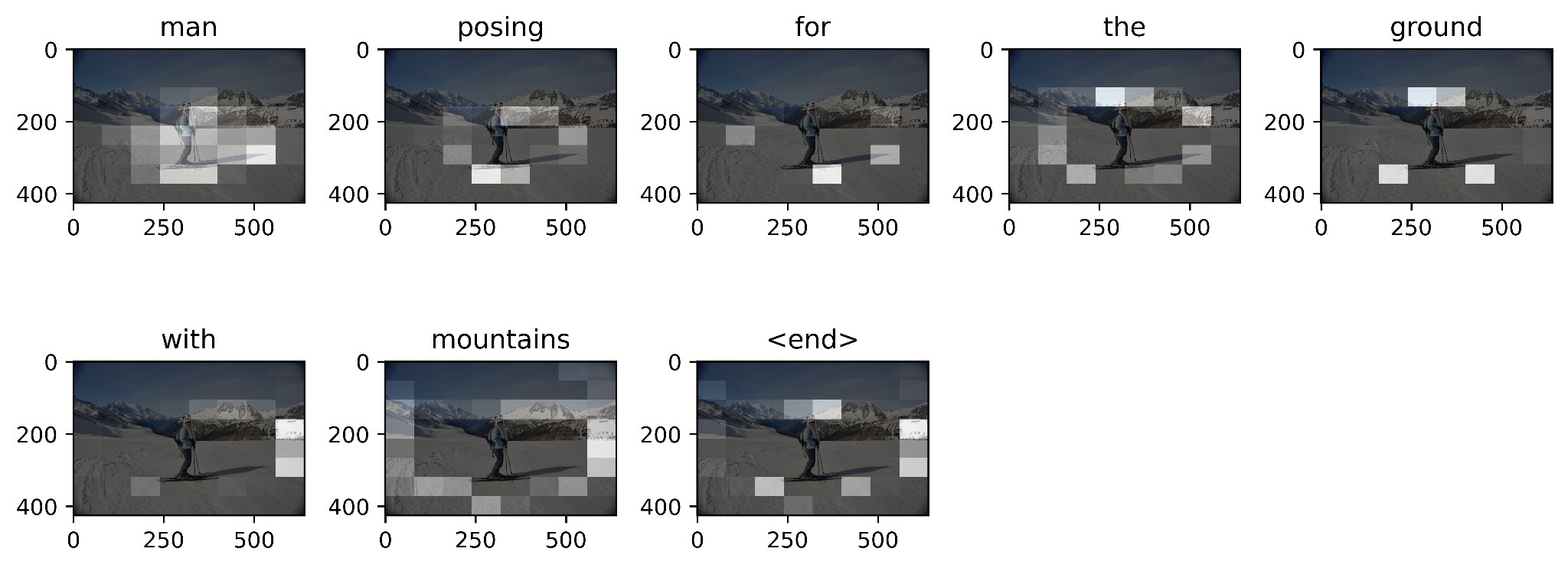}
        \caption{Window sliding across the image and generating captions}
    \label{fig:example}
\end{figure}

\section{Evaluation Methods}
A variety of performance metrics exist for evaluating the efficacy of generated captions. In this study, we employ the BLEU metric [7] to gauge the quality of the captions generated. BLEU evaluates the similarity between the machine-generated text and the reference text through the calculation of n-gram overlap, with higher scores indicating a closer match. BLEU is a commonly used method in the field of machine translation and has applications in other natural language generation tasks as well.

\subsection{METEOR}
In the field of image captioning, METEOR [6] is a widely used evaluation metric that measures the correspondence between words in the generated captions and the reference captions. It typically employs techniques such as Word-Net or Porter- Stemmer to perform a one-to-one mapping of the words in the captions. This mapping is used to compute an F-score, which represents the overall performance of the generated captions. Despite its popularity, METEOR has seen less use in the last few years, particularly with the rise of deep learning models for natural language generation (NLG).

\subsection{SPICE}
SPICE [14] (Semantic Propositional-based Image Caption Evaluation) is a metric for evaluating the performance of image captioning models. It measures the overlap between the model-generated captions and the human-annotated captions in terms of recall, precision, and F1-score. Recall measures the proportion of relevant information in the human-annotated captions that is present in the model-generated captions. Precision measures the proportion of the model-generated captions that are relevant to the human-annotated captions.The F1-score is a single score that balances both measures by taking the harmonic mean of recall and precision.

\subsection{ROUGE}
ROUGE (Recall-Oriented Understudy for Gisting Evaluation) measures the quality of automatically generated captions by comparing them with ideal captions generated by humans. The ROUGE evaluation package contains multiple measures, including {ROUGE\_N}, {ROUGE\_W}, {ROUGE\_S}, and {ROUGE\_L}.

\section{Dataset}
The selection of the dataset significantly impacts the outcome of a deep learning model in image captioning. There are various datasets available for this task, such as the widely-used standard benchmark datasets such as MS COCO [13], Flickr 8k [12]  , and Flickr 30k [11], each of which associates multiple captions with each image.This big data requires optimization to reduce latency and data redundency [23] .The results obtained from different datasets can vary. A significant number of generic photos are needed in a dataset for image captioning, and preprocessing might also be necessary. However, because it does not require preprocessing, using subword-based tokenization approaches, such BPE, is frequently favoured.The quality of the captions and the frequency of the words play a critical role in determining the quality of the dataset. For instance, if an image has multiple captions each with a unique vocabulary, the model may not be able to accurately predict the words to describe the image. Researchers frequently utilise the Karpathy split because it streamlines the evaluation process, allocating 5000 photos for the test set, 5000 images for the validation set, and the remaining 5000 images for training. The table below summarises the size and domain of various datasets available for generic, news, or fashion purposes.

\begin{table}[htbp]
  \centering
  \caption{Dataset Overview}
  \label{tab:example}
  \begin{tabular}{cccccc}
    \hline
    Name &  Domain & Images & Vocab Size & No.Captions \\
    \hline
    COCO & Generic & 132K & 27K(10K) & 5 \\
    Flickr8K & Generic & 8K & 8K(3K) & 5 \\
    Flickr30K & Generic & 30K & 18K(7K) & 5 \\
    Oxford-102 & Flower & 8K & 5K(2K) & 10 \\
    Fasion Caps & Fashion & 130K & 17K(16K) & 1 \\
    Text Caps & OCR & 28K & 44K(13K)& 5\\
    \hline
  \end{tabular}
\end{table}

\begin{figure}
    \centering
        \includegraphics[width=6cm,height=8cm]{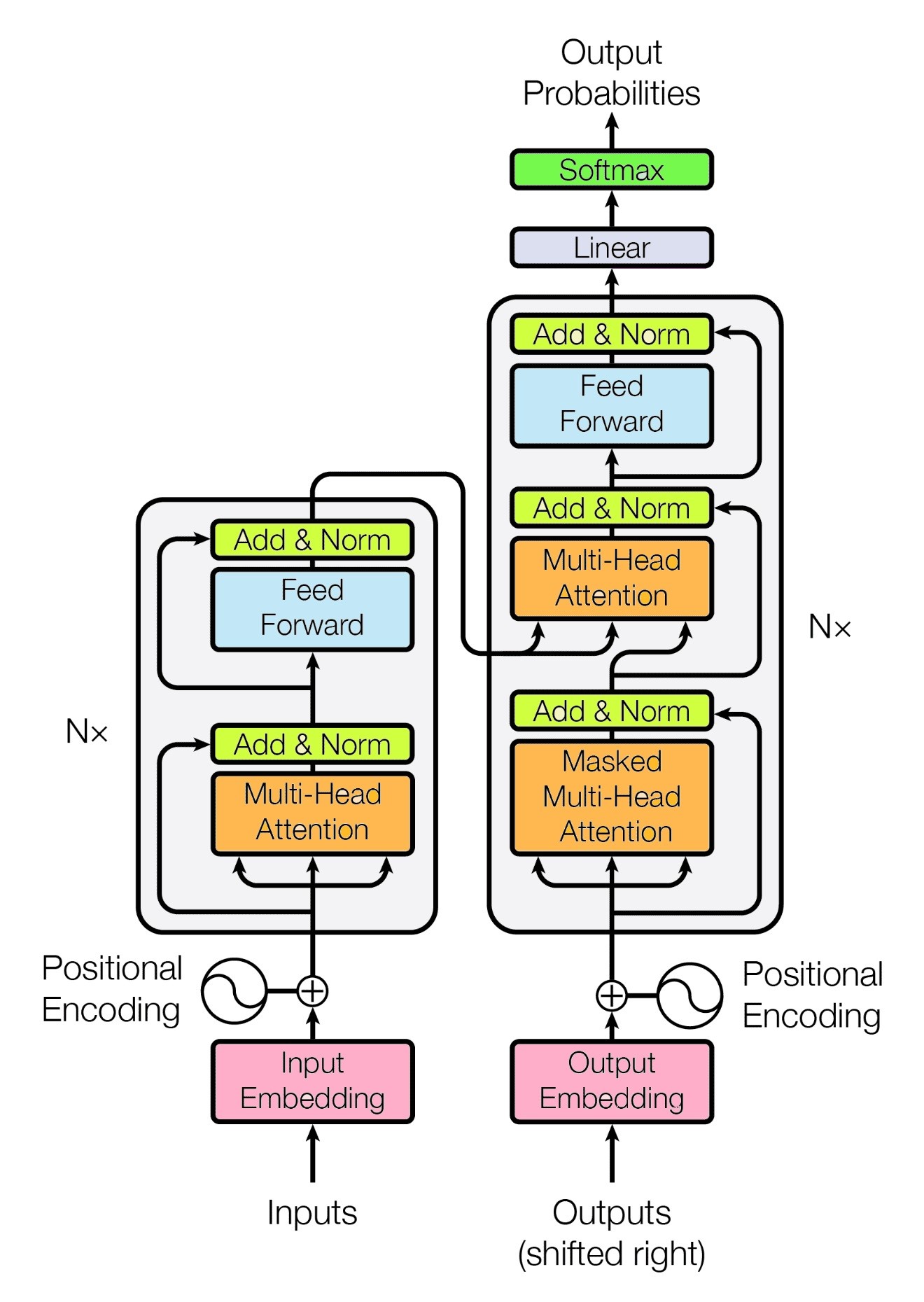}
        \caption{Transformer architecture.}
    \label{fig:example}
\end{figure}

\section{ Proposed Methodology}
In this study we compared two well-known Image captioning models: an LSTM with an Attention block and a Transformer model. The models were trained using the MS COCO-14 dataset, which includes approximately 86,000 training images and 40,000 validation images. The training process was conducted on a g4dn.xlarge GPU instance from AWS, which is equipped with 4 vCPUs and 16 GiB of memory. The entire implementation of both models was done using the Tensorflow platform.

\subsection{Model Architecture}
Researchers have been keen upon finding the best Image Captioning models . Various CNN models were introduced to accomplish this task like the VGG16 [17], VGG19 [17],Inception V3 [16]  ,ResNEt50 [15] . Most of these architectures consisted of  parallel convolutional layers stacked upon each other .Among these different layers , Inception V3 has been one of the most popular choices mainly because of its higher efficiency , is faster and is computationally less expensive . It consists of 42 convolutional  layers and a lower rate of error than its predecessors.

\subsubsection{Inception V3}
Inception deep convolution approach was presented as GoogleNet.This architecture breaks down the convolutions by replacing the existing filters by 1-D filters . The main feature of this  architecture  is that it concatenates multiple different sized convolutional filters into one filter. This reduces the computational complexity as a lesser number of parameters have to be trained. In our model, the image is first passed to InceptionV3, which extracts its characteristics and saves them in a numpy file. Once features are extracted it is passed to the transformer or LSTM network for caption generation tasks.

\subsubsection{Transformer}
Transformer [3] addresses the issue of the vanishing gradient problem that occurs in traditional RNN and LSTM models when processing lengthy text inputs. The attention mechanism, which calculates the importance of each input text, is utilised to overcome this problem.
\newline
\newline
The Encoder and Decoder blocks make up the Transformer, a fully attentive architecture. These blocks are made up of numerous layers of feed-forward, positional encoding, and multi-head attention layers.
\newline
The input, which is produced by combining input embedding and position encoding, is sent into the encoder, which then creates an attended representation with contextual information for each word in the input sentence.
\newline
\newline
A masked multi-head attention layer in the Decoder, in addition to the layers in the Encoder, makes sure that future words are not taken into account when making predictions. The output of the Encoder serves as the input for the multi-head attention layer in the decoder, which generates one word at a time and pays attention to previously generated words.

\subsubsection{Multihead Attention}
It allows the model to attend to different positions in the input sequence, in parallel, allowing for greater representation power.
In Multi-Head Attention, the input sequence is first linearly projected into multiple heads, with each head having its own set of weight matrices. These projections are then processed using scaled dot-product attention to compute attention scores between each query and each key. These scores are then used to compute a weighted sum over the values, which represent the context vectors. Finally, the outputs of each head are concatenated and linearly transformed back to the original representation space.
\newline
A stronger representation of the input is produced because of the Multi-Head Attention mechanism, which allows the Transformer to process relationships between several points in the input sequence simultaneously and pay attention to the most crucial information in each head.

\begin{figure}
    \centering
        \includegraphics[width=5cm,height=6cm]{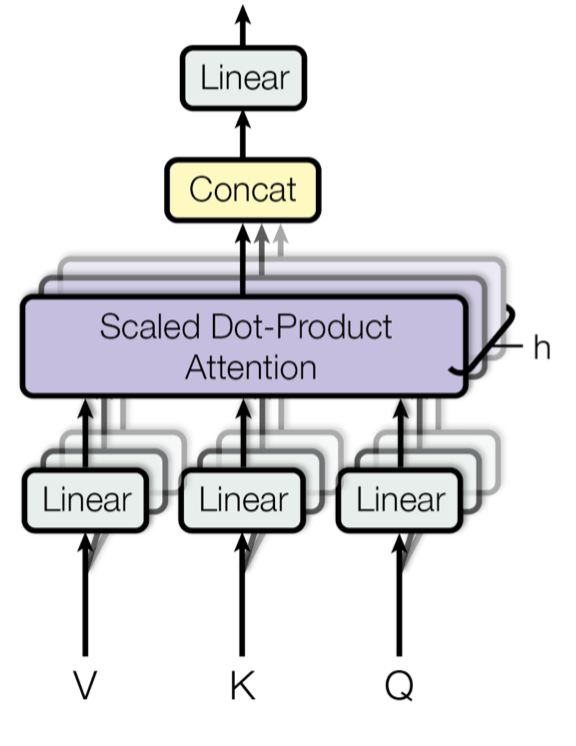}
        \caption{Multihead attention}
    \label{fig:example}
\end{figure}

\subsubsection{LSTM}
Researchers frequently employ LSTM (Long Short Term Memory) [18] in text translation, audio to text conversion, and other applications. Its core architecture is comparable to that of RNNs, but its repeating module design is different.It has basic architecture similar to RNNs but differs in the design of iteration modules .It has structures called as cell states which preserves selective  recent information in it . The information to be preserved passes through structures called gates . 
\newline
\newline
Gates  decide which part of the information should be stored and which part of the information should be discarded .The sigmoid functions that make up these gates have values between 0 and 1. The information is sent on if the value returned is 1, otherwise it is discarded.

\begin{figure}
    \centering
        \includegraphics[width=4cm,height=4cm]{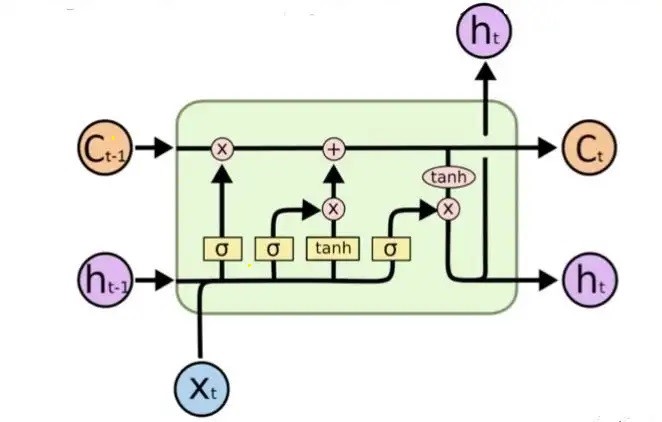}
        \caption{LSTM architecture}
    \label{fig:example}
\end{figure}

\subsubsection{Attention Mechanism in LSTM}
While making a prediction, the Attention Mechanism in LSTMs [10] is a method used to direct the network's attention on a specific portion of the input sequence. It assists the model in image captioning to concentrate on the focal points of a picture and provide a relevant caption.
\newline
An LSTM with Attention processes the input sequence to create hidden states, which are then fed via an attention mechanism to calculate attention scores. The hidden states are weighted using these scores to produce a context vector that reflects the most crucial data in the input sequence. The final prediction is then created by concatenating the context vector with the LSTM's hidden state and passing it through a fully connected layer.
The Attention Mechanism in LSTMs provides the network with the ability to focus on specific parts of the input and attend to the most relevant information when generating predictions, leading to improved performance in tasks such as Image Captioning. Additionally, it helps the model to overcome the vanishing gradient problem commonly seen in longer sequences and allows the network to generate captions that are more relevant to the image.

\begin{figure}
    \centering
        \includegraphics[width=4cm,height=5cm]{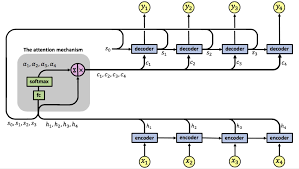}
        \caption{Attention block in Encoder Decoder architecture.}
    \label{fig:example}
\end{figure}

\subsubsection{BLEU Score}
A well-liked metric for assessing the quality of machine-generated text in comparison to a reference corpus is BLEU (Bilingual Evaluation Understudy) [7]. The score is determined by how many n-grams in the generated text and reference text overlap, where n can range from 1 to 4. The BLEU score increases with increasing overlap.
\newline
\newline
In our study, the BLEU score was used to assess how well our text generation model performed. The generated texts were created by the model, and the reference corpus was made up of a collection of human-written texts. To determine an overall performance metric, the BLEU score was calculated for each generated text and averaged throughout the full reference corpus.

\begin{figure}
    \centering
        \includegraphics[width=6cm,height=5cm]{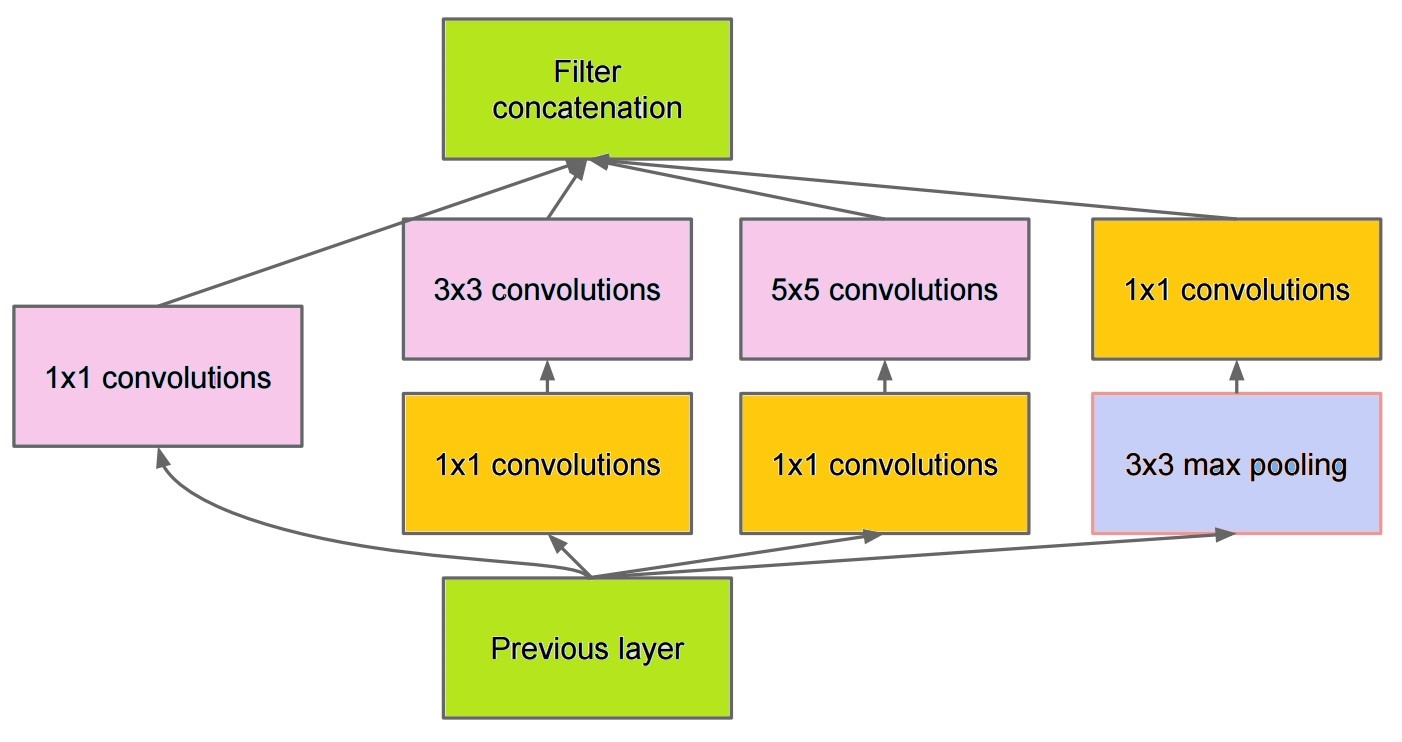}
        \caption{InceptionV3 architecture.}
    \label{fig:example}
\end{figure}

\section{Results}
In this study, two models were trained using the same dataset and under comparable conditions. The results indicated that the Transformer model outperformed the LSTM with Attention block model in terms of accuracy. However, in the early stages of training, both models displayed low accuracy and produced non-relevant captions. The Transformer model was initially trained for just 10 epochs to evaluate its performance in comparison to the other model. The results were remarkable, with a Train Accuracy of 51.22\%, Validation Accuracy of 48.80\%, a Training loss of 11.731\%, and a Validation loss of 13.1320\%. The best BLEU score achieved was 0.45, while the best BLEU score we got for the LSTM model was 0.3.

\begin{figure}
    \centering
        \includegraphics[width=5cm,height=6cm]{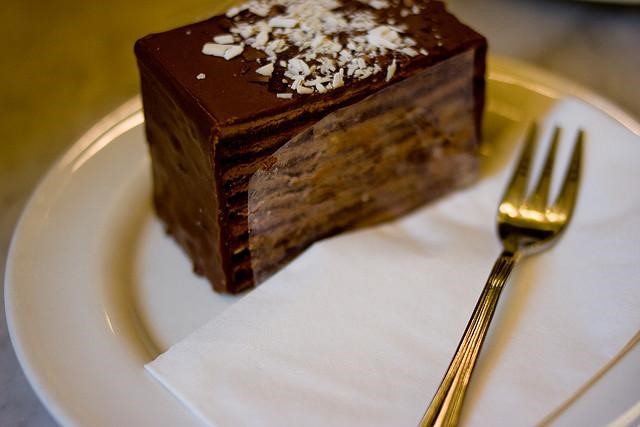}
        \caption{Predicted caption : Chocolate cake inside of a display on it a table <end>.}
    \label{fig:example}
\end{figure}

\section{Discussion}
In this research paper, we seek to gain a fundamental understanding of image captioning and to conduct a comparative study of various image captioning models. Numerous architectures have produced state-of-the-art results in image caption generation. In this study we have generated captions using a base transformer architecture and achieved a validation accuracy of 48.8\% after training for 10 epochs and base LSTM network with attention mechanism on the standard MS COCO dataset, which consists of around 86,000 training images and 6000 validation images. There have been several modifications made on the transformer architecture that have resulted in state-of-the-art accuracy, such as the Meshed Memory Transformer [20] and X Linear Attention [19]. The Clipped Diffusion Model [4]  has also demonstrated an improvement in accuracy.

\subsubsection{M2 Transformer}
Combine a multi-layer encoder for image areas with a multi-layer decoder to produce the output sentence. Encoding and decoding layers are connected in a mesh-like structure, weighted by a learnable gating mechanism, to take advantage of both low-level and high-level contributions. Relationships between image regions are encoded in a multi-level manner by utilising learned a priori knowledge, which is modelled via persistent memory vectors. Generated BLEU1 score of 81.6.

\subsubsection{X Linear Attention}
 A variation of the self-attention mechanism utilised in image captioning models is the X-Linear attention mechanism. The goal of the X-Linear attention mechanism is to increase the effectiveness of the attention mechanism in models for image captioning.The X-Linear attention mechanism uses a linear transformation to compute the attention scores instead of the dot product. This reduces the computational complexity and memory requirements of the attention mechanism, making it more efficient. Additionally, the linear transformation
 can be implemented using low-dimensional embeddings. Obtained new state-of-the-
art performances on this captioning dataset with X-LAN.

\subsubsection{Clip Diffusion Model}
CLIP model is a contrastive-learning-based model trained on WebImageText dataset. WebImageText consists of 400 million image-text pairs collected from publicly available sources on the Internet. The CLIP model demonstrates strong zero-shot performance in their evaluation on the ImageNet  dataset. Further work on the CLIP model also shows its transferability to other tasks.
\newline
Diffusion is a generative model which generates new data. The idea behind a diffusion model is to diffuse information through a continuous latent space over time. Diffusion models have been used in various tasks like image generation, language generation tasks. 
This study successfully applied diffusion probabilistic model to image caption generation task. On Flickr8k and Flickr30k combined dataset this model achieved the BLEU4 score of 0.247.

\begin{table}[htbp]
  \centering
  \caption{Different models performance}
  \label{tab:example}
  \begin{tabular}{cccccc}
    \hline
    Model & B1  & B4 & M & C \\
    \hline
    Up-Down (ResNet-101) & 80.2 & 36.9 & 27.6 & 117.9 \\
    SGAE (ResNet-101) & 81.0 & 38.5 & 28.2 & 123.8 \\
    X-LAN (ResNet-101) & 81.1 & 39.5 & 29.4 & 128.0 \\
    X-Transformer (ResNet-101) & 81.3 & 39.9 & 29.5 & 129.3 \\
    M2 Transformer & 80.8 & 39.1 & 29.2 & 131.2 \\
    AOA-NET    & 80.2 & 38.9 & 29.2 & 129.8\\
    \hline
  \end{tabular}
\end{table}

\section{Conclusion}
Two different architectures for Image caption generation were compared on the basis of their performance . Inception V3 CNN encoders used in both the models for feature extraction of image .The dataset used for this study  is MS COCO 2014 which consists of 86000 train images and 40000 validation  images . g4dn.xlarge , a GPU instance on AWS was used to process this huge dataset .We can conclude that the Transformer model achieved a better accuracy and BLEU score than the Long Short Term Memory  model with Attention block .Although Transformer models are known for their accuracy , we have some other models which are relatively new and have shown state of the art accuracy  in the task of Image captioning . These are the CLIP-diffusion model , X - Linear Attention model and M2 transformer.

\end{document}